\documentclass[letterpaper,twocolumn,fleqn]{article} 

\usepackage{ist}
\usepackage{amssymb}
\usepackage{amsmath}
\usepackage{graphicx}
\usepackage{hyperref}
\usepackage{multirow}
\usepackage[table]{xcolor}
\title{Prune the Convolutional Neural Networks with Sparse Shrink}
\author{Xin Li,
Changsong Liu \\
State Key Laboratory of Intelligent Technology and Systems,\\
Tsinghua National Laboratory for Information Science and Technology,\\
Department of Electronic Engineering, Tsinghua University, Beijing 100084, China\\
Email: \{lixin08, lcs\}@ocrserv.ee.tsinghua.edu.cn
}

\date{} 

\begin{document} 

\maketitle 

\thispagestyle{empty} 


\begin{abstract}
Nowadays, it is still difficult to adapt Convolutional Neural Network (CNN) based models for deployment on embedded devices. The heavy computation and large memory footprint of CNN models become the main burden in real application. 
In this paper, we propose a ``Sparse Shrink'' algorithm to prune an existing CNN model. 
By analyzing the importance of each channel via sparse reconstruction, the algorithm is able to prune redundant feature maps accordingly.
The resulting pruned model thus directly saves computational resource. 
We have evaluated our algorithm on CIFAR-100. As shown in our experiments, we can reduce $56.77\%$ parameters and $73.84\%$ multiplication in total with only minor decrease in accuracy. These results have demonstrated the effectiveness of our ``Sparse Shrink'' algorithm.
\end{abstract}

\section{Introduction}
\label{sec:intro}
In recent years,  great progress has been achieved in computer vision which is arguably attributed to greater computation resources and the application of deep learning algorithms \cite{szegedy2015going,simonyan2014very,long2015fully,ren2015faster}. The convolutional neural networks (CNN) is a popular example of deep learning algorithms. It adopts a deep architecture that consist of many stacked convolutional and fully-connected layers, which is specifically designed for solving computer vision related problems. Although CNN has bring breakthrough into computer vision, we are still not possible to decide the optimum network architecture, \emph{e.g.} number of channels in convolutional layer, for a specific task. Nowadays, people tend to design large networks with large number of channels to build a high-capacity model. However, this brings a large demand on computation and memory capacity, which are especially limited on embedded devices. The heavy computation and large memory footprint of CNN models become the major burden in real application. 

On the other hand, it is observed that there is redundancy in large networks \cite{han2015learning,wen2016learning}. 
Convolutional layers occupy the main calculation in CNN, and the responses of their resulting feature maps are sometimes largely correlated to each other. 
Therefore, it is intuitive to prune a large pre-trained model by removing redundant connections. This will results in a lightweight network with comparable level of performance and less demand on both memory and computational complexity.

Motivated by this, we propose a novel ``Sparse Shrink'' algorithm to prune a CNN model: we evaluate the importance of each channel of feature maps, and prune less important channels to get a slimmer network. The pruned model is of a similar performance with original model, yet thinner structure and lower computational complexity.

\begin{figure}[!tb]
  \includegraphics[width=0.95\columnwidth]{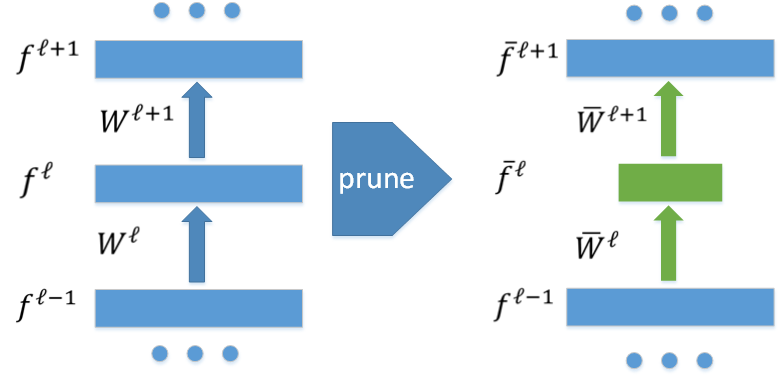}
  \caption{By evaluating the importance of each channel, ``Sparse Shrink'' prunes less important channels and builds a slimmer model. Weights in the upper and lower layer are modified accordingly.}
  \label{fig:demo}
\end{figure}

\section{Related Work}
Extensive work have been done to accelerate the testing of
CNN models or lower its memory cost. Some \cite{jaderberg2014speeding,zhang2015efficient} of them
speed up the testing by explore the sparsity in CNN models with
low rank decomposition.
Vasilache \cite{vasilache2014fast} speed up the convolution
operation by a Fast Fourier Transform implementation. However,
these algorithms focus on either accelerating test speed or
lower memory footprint of CNN models without changing their
model structures. 

Network pruning has been studied by several researchers \cite{lecun1989optimal, hassibi1993second,stepniewski1997pruning,rastegari2016xnor} . Lecun \emph{et al.} \cite{lecun1989optimal} and Hassibi \emph{et al.} \cite{hassibi1993second} show that a portion of weights can be set to zero by analyzing their value and Hessian matrix. 
Han \emph{et al.} \cite{han2015learning,han2015deep} gradually prune the small-weights in a network, and further reduce storage requirement by compressing weights in fully connected layer with matrix factorization and
vector quantization.
Rastegari \emph{et al.} \cite{rastegari2016xnor} binarize both the weights and layer inputs, such that the resulting network mainly uses XNOR operations.
Stepniewski \emph{et al.} \cite{stepniewski1997pruning} prunes network with genetic algorithm and simulated annealing.
However, these algorithms only makes use of intra-kernel sparsity, without doing channel wise pruning. This limits GPUs to expolit computational savings.
Different from existing algorithms, our ``Sparse Shrink'' algorithm directly prune network structure in convolutional layer by channel wise pruning. The most related work on channel wise pruning would be ``Structured pruning'' \cite{anwar2015structured}. It naively remove the incoming and outgoing weights of a pruned channel. In contrast, we modify convolutional kernel in the upper layer by reconstructing original feature maps in order to reduce  decrease in accuracy.

\begin{figure*}[!t]
  \centering
  \includegraphics[width=0.95\linewidth]{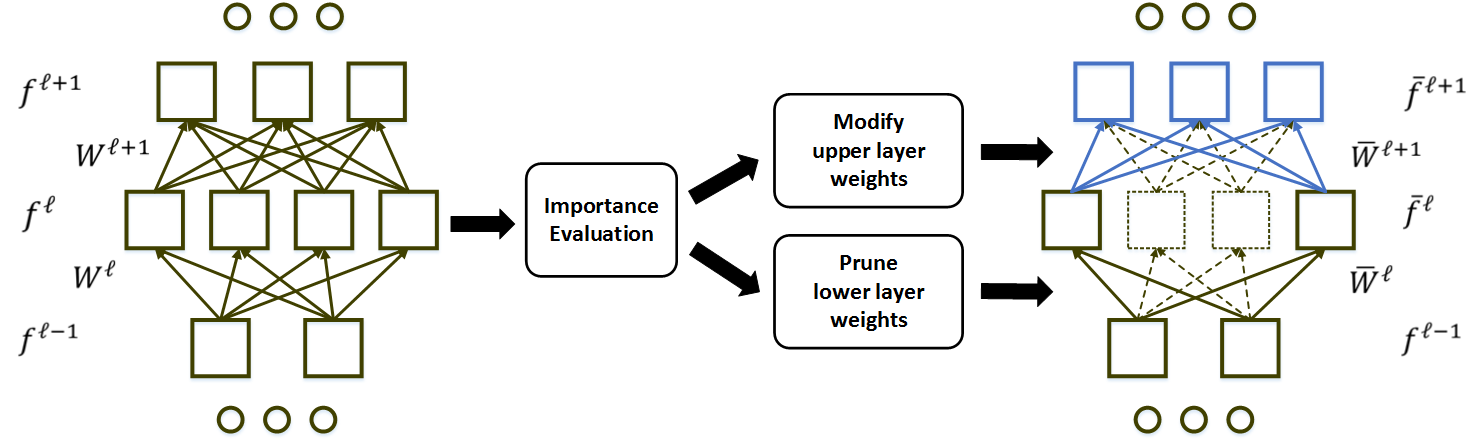}
  \caption{Illustration of ``Sparse Shrink'' algorithm. We evaluate the importance factor of each channel of feature maps $f^\ell$, and prune the least important channels (dashed box). The pruning operation involves removing corresponding channels in $W^\ell$ (dashed line), and modifying  convolutional kernel $\overline{W}^\ell$ (blue line).}
  \label{fig:pipeline}
\end{figure*}

\section{Sparse Shrink}
In this section, we elaborate how our ``Sparse Shrink'' algorithm prune an existing network by channel-level pruning in convolutional layer. The basic idea of ``Sparse Shrink'' is intuitive: there exists redundancy in convolutional layers, and we can remove redundant channels to produce a pruned model with minimum loss in accuracy. 
There are three major steps in our algorithm. Firstly, we evaluate the importance of each channel with ``Sparse Reconstruction'' algorithm. 
Secondly, those redundant, \emph{i.e.} less important channels, are removed, and related convolutonal kernels are modified, as shown in Figure \ref{fig:pipeline}.
This results in a pruned model with a minor decrease in accuracy. 
Finally, the pruned model is re-trained to achieve its best performance.

\subsection{Importance Evaluation}
Sparse reconstruction \cite{elhamifar2012see,mairal2008discriminative,ramirez2010classification} is a well-studied problem which focus on finding representative data points, such that each data point in the dataset can be described as a linear combination of a set of representative points.
Formally, with a data matrix $D\in \mathbb{R}^{m\times N}$, \emph{i.e.} $N$ data points of a dataset in $\mathbb{R}^m$, the standard $\ell_{1}$ relaxation of the optimization problem can be written as
\begin{equation}
\label{eqn:sparse_reconstruction}
min\left \| D-DU \right \|^{2}_{F}, s.t. \left \| U \right \|_{1,q}\leq \tau, \mathbf{1}^{\top }\mathbf{U}=\mathbf{1}^{\top }
\end{equation}
where $U\in \mathbb{R}^{N\times N}$ is corresponding reconstruction coefficient matrix and 
$\left \| U \right \|_{1,q} \triangleq \sum_{i=1}^{N} \left \| u^i \right \|_{q}$ 
 is the sum of the $\ell_q$ norms of the rows of $U$. We choose $q=2$ so that the optimization program is convex and $\tau>0$ is an appropriately chosen parameter. $\mathbf{1}^{\top }\mathbf{U}=\mathbf{1}^{\top }$ is a affine constraint to make the representatives be invariant with respect to a global translation of the data.

Now we elaborate how to make use of sparse reconstruction to evaluate the importance of each channel in a convolutional layer. Throughout this paper, we use the following notations for the simplicity of explanation. Let $f^\ell$ denote the output feature maps for the $\ell$-th layer and $f^\ell_i$ denote the value of the $i$-th channel. The feature maps has a dimension of $C_\ell \times H\times W$, where $C_\ell$ is the number of channels in layer $\ell$, and $H \times W$ is the corresponding spatial size. To evaluate the importance of each channel in feature maps $f^\ell$, we randomly select $N$ input image, and get a data matrix $D^{N\times C_\ell \times H\times W}$. 
In contrast to standard sparse reconstruction algorithm as Equation \eqref{eqn:sparse_reconstruction}, which focus on finding representative data points among $N$ total data points, our algorithm aims at finding representative channels among the $C_\ell$ channels. Therefore we reshape the data matrix into $D^{ \left ( N\times H\times W \right ) \times C_\ell}$, and regard each channel $c_i$ as a ``data point'' in $\mathbb{R}^{N\times H\times W}$. With this representation, we are able to find the most representative channels by reconstructing data matrix $D$. 

More specifically, we use the entire data matrix as dictionary and try to reconstruct the data matrix with reconstruction coefficients $U \in \mathbb{R}^{C_\ell \times C_\ell}$.
\begin{eqnarray}
\begin{bmatrix} d_1 & d_2 & ... & d_{C_\ell} \end{bmatrix} \approx \begin{bmatrix} d_1 & d_2 & ... & d_{C_\ell}  \end{bmatrix}\begin{bmatrix} u^1\\ u^2\\ ...\\ u^{C_\ell} \end{bmatrix} \nonumber
\end{eqnarray}
Then we solve the optimization problem in Equation \eqref{eqn:sparse_reconstruction} to get the reconstruction coefficients $U$. 

The regularization term $\left \| U \right \|_{1,2} \triangleq \sum_{i=1}^{C_\ell} \left \| u^i \right \|_{2}$ in Equation \eqref{eqn:sparse_reconstruction} 
provides information about relative importance between channels. A more representative channel takes larger part in reconstruction, and thus the corresponding reconstruction coefficients have more non-zeros elements with larger values. Hence, the resulting coefficients can be intuitively utilized to rank importance of each channel, and to evaluate feature maps redundancy. More precisely, we rank a channel $i$ by its importance factor $\left \| u^i \right \|_{2}$, where $u^i \in \mathbb{R}^{1\times C_\ell}$ indicates the $i$-th row of reconstruction matrix $U$. The lower importance factor is, the more redundant the channel become. Therefore, we prune these bottom-ranking channels to get a slimmer network.

\subsection{Network Pruning}
Once we rank the importance factors, we can prune the network in layer $\ell$, by removing the least important $K$ channels.
This involves two specific modifications in network weights, removing channels in layer $\ell$ and reconstructing feature maps in layer $\ell+1$.

As illustrated in Figure \ref{fig:pipeline}, the feature maps $f^{\ell}$ are obtained by convolving $f^{\ell-1}$ with kernel $W^\ell \in \mathbb{R}^{C_\ell \times C_{\ell-1} \times k\times k}$, where $k$ is the spatial size of convolutional kernel. To remove a channel $c_i$ in $f^{\ell}$, we only need to remove corresponding ``Slice'' in $W^\ell$, \emph{i.e.} $W^{\ell}_{c_i} \in \mathbb{R}^{C_{\ell-1} \times k\times k}$. 
Having pruned $K$ least important feature maps, the new pruned convolutional kernel  $\overline{W}^{\ell} \in \mathbb{R}^{\left ( C_{\ell}-K \right ) \times C_{\ell-1}\times k\times k}$ has a channel number $C_{\ell}-K$. And the new feature maps $\overline{f^{\ell}} \in \mathbb{R}^{\left ( C_{\ell}-K \right ) \times C_{\ell-1}\times H\times W}$ is obtained by convolving $\overline{W}^{\ell}$ with $f^\ell$.

Pruning layer $\ell$ will obviously affect layer $\ell+1$. 
Instead of naively removing corresponding channels in $W^{\ell+1}$, we manage to get a new convolutional kernel by reconstructing the original feature maps $f^{\ell}$, in order to minimize the decrease in accuracy after pruning. Given a data matrix $\overline{D} \in \mathbb{R}^{\left ( C_{\ell}-K \right ) \times\left ( N\times H\times W \right )}$ of pruned feature maps $\overline{f^\ell}$,
we try to reconstruct original $f^{\ell}$ data matrix by minimizing reconstruction error,
\begin{eqnarray}
\label{eqn:reconstruction}
\min Err & = & \min_{V}\left \| D-\overline{D}V \right \|
\end{eqnarray}
Where $V \in \mathbb{R}^{\left ( C_\ell-K \right )\times C_\ell}$ is the reconstruction coefficients.
We can obtain a closed-form solution for Equation \eqref{eqn:reconstruction},
\begin{eqnarray}
\label{eqn:reconstruction_weights}
V & = & \left ( \overline{D}^{\top }\overline{D} \right )^{-1}\overline{D}D
\end{eqnarray}
Let $\widehat{V} \in \mathbb{R}^{C_\ell \times \left ( C_\ell-K \right ) \times 1 \times 1}$ denote the $1\times1$ convolutional kernel derived from $V$, where $\widehat{V}_{i,j,1,1}\triangleq V_{j,i}$. The reconstructed feature maps $\widehat{f}^\ell$ is obtained with,
\begin{eqnarray}
\widehat{f}^\ell =  \widehat{V} \ast \overline{f^{\ell}} \nonumber
\end{eqnarray}
And the feature maps $\overline{f}^{\ell+1}$ in the pruned network can thus be written as,
\begin{eqnarray}
\overline{f}^{\ell+1} & = & ReLU \left (  W^{\ell+1} \ast \widehat{f}^\ell \right ) \nonumber\\
                              & = &  ReLU \left (  W^{\ell+1} \ast \left ( \widehat{V} \ast \overline{f^{\ell}} \right ) \right ) \nonumber\\
                              & = & ReLU \left ( \left ( W^{\ell+1} \ast  \widehat{V} \right ) \ast \overline{f^{\ell}} \right ) \nonumber
\end{eqnarray}
And the new convolution kernel $\overline{W}^{\ell+1} \in \mathbb{R}^{C_{\ell+1}\times \left ( C_\ell-K \right )}$ is,
\begin{eqnarray}
\overline{W}^{\ell+1} &=&  W^{\ell+1} \ast  \widehat{V} \nonumber\\
 &=&  W^{\ell+1} V^{\top}
\end{eqnarray}

Now we get a pruned network with $C_\ell-K$ channels in layer $\ell$, and pruned convolution kernels  $\overline{W}^{\ell}$,  $\overline{W}^{\ell+1}$. The newly pruned model may perform better after further training for more iterations.

\section{Experiment}
We evaluated the performance of ``Sparse Shrink'' algorithm
on the benchmark dataset CIFAR-100 \cite{krizhevsky2009learning}. CIFAR-100 is a widely used benchmark dataset for image classification with $60,000$ color images of 100 categories in total. This size of images is $32\times32$. Images are split into $50,000$ training set and $10,000$ test set. Following NIN \cite{lin2013network} we use global contrast normalization and ZCA whitening as pre-processing.
We use NIN \cite{lin2013network} model as a baseline model, which has been proven to be a successful CNN structure on CIFAR-100.
There are three convolutional layers in the NIN model, \emph{i.e.} $Conv1$,$Conv2$,$Conv3$, with $192$ channels in each of them. In this paper we focus on pruning these three convolutional layers to obtain slimmer networks.
We employ Caffe \cite{jia2014caffe} implementation as our experiment platform. Throughout the experiments, we fix the initial learning rate to $0.01$ and the weight decay coefficient to $0.001$. The code and models is released at: \href{url}{https://github.com/lixincn2015}.

\begin{figure}[!tb]
  \includegraphics[width=0.9\columnwidth]{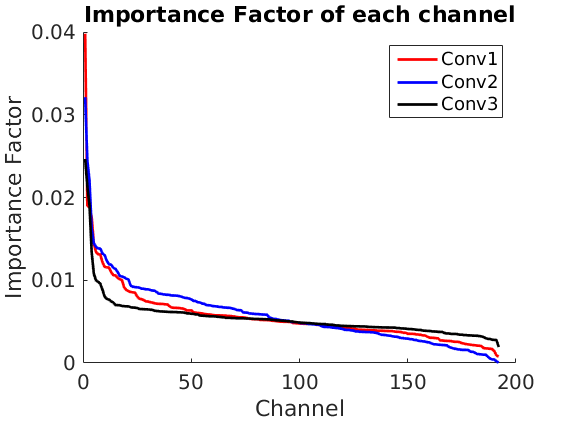}
  \caption{Importance factors of each channel in the three convolutional layer.}
  \label{fig:compareC}
\end{figure}

\begin{table}[!t]
\label{tab:cifar}
\centering
\caption{Table 1. Comparison of accuracies (\%) of the pruned models on CIFAR-100 test set.}
\hspace{3cm}
\renewcommand{\arraystretch}{1.3}
\begin{tabular}{c|cccccc}
\hline
Pruned                      & \multirow{2}{*}{0} & \multirow{2}{*}{64} & \multirow{2}{*}{96} & \multirow{2}{*}{128} & \multirow{2}{*}{160} & \multirow{2}{*}{176} \\
\multicolumn{1}{l|}{Channels} &                    &                     &                     &                      &                      &                      \\ \hline
Conv1                         & 68.08              & 67.80               & 67.86               & 67.86                & 67.36                & \textcolor{blue}{67.38}                \\
Conv2                         & 68.08              & 67.51               & 67.36               & \textcolor{blue}{66.98}                & 65.95                & 64.67                \\
Conv3                         & 68.08              & 67.68               & \textcolor{blue}{67.00}               & 66.07             & 65.09                & 61.17                \\ \hline
\end{tabular}
\end{table}


We conduct three sets of experiments to evaluate our algorithm.
In the first experiment, we apply ``Sparse Shrink'' algorithm to each of the three convolutional layers separately. 
And the sorted importance factors of each layer are shown in Figure \ref{fig:compareC}.
As shown in Figure \ref{fig:compareC}, there are some channels with obviously larger importance in all three convolutional layers, while others have relatively smaller ones. Pruning those channels with smaller importance factors is supposed to result in less decrease in performance.

By pruning different number of channels according to importance factors, we get corresponding pruned models and then evaluate these models on CIFAR-100 test set. Detailed result is shown in Table 1 \ref{tab:cifar}, where $Conv1$,$Conv2$,$Conv3$ are three convolutional layers from the bottom up. The baseline NIN model, \emph{i.e.} not pruning any channels on any layer, has an accuracy of $68.08\%$. 
As shown in Table 1 \ref{tab:cifar}, with a decrease of $\sim1\%$ in accuracy, we can prune as many as $176$, $128$, and $96$ channels on three convolutional layers respectively (highlighted in blue). It is worth mentioning that pruning $176$ channels on $Conv1$ layer brings only minor decrease of  $0.7\%$  in accuracy. We attribute this to the effectiveness of our ``Sparse Shrink'' algorithm, which can dramatically reduce redundancy in feature maps while preserving important information. 

\begin{table*}[!thb]
\centering
\caption{Table 2. Comparison of number of parameters and multiplication between pruned model and baseline model.}
\hspace{1.5cm}
\label{tab:compare_shrunk}
\renewcommand{\arraystretch}{1.3}
\begin{tabular}{c|c|ccc|ccc}
\hline
\multirow{2}{*}{Layer} & \multirow{2}{*}{Input Size} & \multicolumn{3}{c|}{Number of Parameters} & \multicolumn{3}{c}{Number of Multiplications} \\ \cline{3-8} 
 &  & Baseline Model & pruned model & Reduction (\%) & Baseline Model & pruned model & Reduction (\%) \\ \hline
Conv1 & $32 \times 32$ & $192 \times 3 \times 5 \times 5$ & $16 \times 3 \times 5 \times 5$ & 91.67 & $1.47 \times 10^7$ & $1.23 \times 10^6$ & 91.67 \\
Cccp1 & $32 \times 32$ & $160 \times 192 \times 1 \times 1$ & $160 \times 16 \times 1 \times 1$ & 91.67 & $3.15 \times 10^7$ & $2.62 \times 10^6$ & 91.67 \\
Cccp2 & $32 \times 32$ & $96 \times 160 \times 1 \times 1$ & $96 \times 160 \times 1 \times 1$ & 0 & 1$.57 \times 10^7$ & $1.57 \times 10^7$ & 0 \\
Conv2 & $16 \times 16$ & $192 \times 96 \times 5 \times 5$ & $64 \times 96 \times 5 \times 5$ & 66.67 & $1.18 \times 10^8$ & $3.93 \times 10^7$ & 66.67 \\
Cccp3 & $16 \times 16$ & $192 \times 192 \times 1 \times 1$ & $192 \times 64 \times 1 \times 1$ & 66.67 & $9.44 \times 10^6$ & $3.15 \times 10^6$ & 66.67 \\
Cccp4 & $16 \times 16$ & $192 \times 192 \times 1 \times 1$ & $192 \times 192 \times 1 \times 1$ & 0 & $9.44 \times 10^6$ & $9.44 \times 10^6$ & 0 \\
Conv3 & $8 \times 8$ & $192 \times 192 \times 3 \times 3$ & $96 \times 192 \times 3 \times 3$ & 50.00 & $2.12 \times 10^7$ & $1.06 \times 10^7$ & 50.00 \\
Cccp5 & $8 \times 8$ & $192 \times 192 \times 1 \times 1$ & $192 \times 96 \times 1 \times 1$ & 50.00 & $2.36 \times 10^6$ & $1.18 \times 10^6$ & 50.00 \\
Cccp6 & $8 \times 8$ & $100 \times 192 \times 1 \times 1$ & $100 \times 192 \times 1 \times 1$ & 0 & $1.23 \times 10^6$ & $1.23 \times 10^6$ & 0 \\ \hline
Overall & - & $9.83 \times 10^5$ & $4.25 \times 10^5$ & 56.77 & $3.23 \times 10^8$ & $8.45 \times 10^7$ & 73.84 \\ \hline
\end{tabular}
\end{table*}

Pruning any one of three convolutional layer results in decreased performance, wheres the decrease show different features.
Pruning lower layers brings less accuracy decrease. 
More specifically, with the same level of decrease in accuracy (highlighted in blue), we can prune much more channels in $Conv1$ than $Conv3$ (176 vs 96).
It indicates that there is more redundancy in the lower layers of NIN model than in the upper layers, and $Conv1$ needs much less feature maps than $Conv3$. 
This finding is consistent with previous studies \cite{zeiler2014visualizing,simonyan2014very}. It's well observed that there is a hierarchical nature of the features in deep networks. Feature maps of lower layers mostly responds to low-level visual features, \emph{e.g.} edges or corners, which can be shared between high-level patterns. Upper layers then assemble the low-level features to exponentially more complex visual patterns. Hence we need a lot more channels in upper layers than in lower layers.

\begin{figure}[!tb]
  \includegraphics[width=0.9\columnwidth]{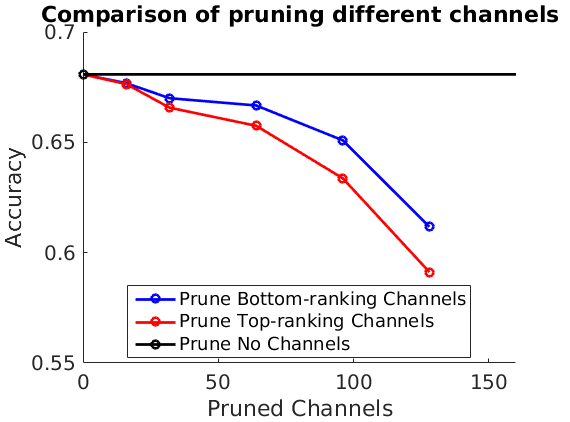}
  \caption{Comparison of pruning top-ranking and bottom-ranking channels in Conv3.}
  \label{fig:prune}
\end{figure}

In the second experiments, we compare the accuracy of pruning different channels in $Conv3$ layer. More specifically, we prune top-ranking and bottom-ranking channels according to importance factors, and evaluate the pruned models on test set. As shown in Figure \ref{fig:prune}, pruning both top-ranking and bottom-ranking channels results in decrease in accuracy. However, pruning bottom-ranking channels brings less decrease. As the number of pruned channels increases, the gap becomes larger. And pruning $128$ bottom-ranking channels has an advantage of $2\%$ over pruning top-ranking channels (61.17\% vs 59.12\%) .
This validates that our ``Sparse Shrink'' algorithm is able to successfully evaluate the importance of each channel, and hence keep the most important feature maps during pruning.

Finally, in the third experiment, we further prune all the three convolutional layers in the network from the bottom up, and remove $176$,  $128$, and $96$ channels in $Conv1$, $Conv2$, $Conv3$ respectively. The final pruned model has an accuracy of $65.53\%$ on test set. Table \ref{tab:compare_shrunk} provides a detailed comparison between baseline model and the pruned model in terms of number of parameters and number of multiplication. For a convolutional kernel $W^\ell \in \mathbb{R}^{C_\ell \times C_{\ell-1} \times k \times k}$ in layer $\ell$, the corresponding number of parameter is $C_\ell \times C_{\ell-1} \times k \times k$. And the number of multiplication in layer $\ell$ is $C_\ell \times C_{\ell-1} \times k \times k \times H \times W$, where $H$ and $W$ are the input size of layer $\ell$.
Compared to the baseline model, this pruned model reduces $56.77\%$ parameters and $73.84\%$ multiplication, at a minor decrease of $2.55\%$ in accuracy. This validates that our ``Sparse Shrink'' algorithm is able to save computational resource of a well-trained model without serious performance degradation. 

\section{Conclusion}
In this paper, we propose a ``Sparse Shrink'' algorithm for convolutional neural network pruning. The Sparse Shrink algorithm evaluates the importance of each channel by sparse reconstruction. Channels with smaller importance factors is considered to be more redundant, and is pruned to get a slimmer network. New convolutional kernels can be derived from reconstructing original feature maps. Experiments on CIFAR-100 dataset show that the ``Sparse Shrink'' algorithm is able to significantly save computational resource with only minor decrease in performance.

\section{Acknowledgments} 
This work was supported by the National Natural Science Foundation of China under Grant No. 61471214 and the National Basic Research Program of China (973 program) under Grant No. 2013CB329403 .

\renewcommand\refname{Reference}
\bibliographystyle{plain}
\bibliography{sparse_shrink}





\end{document}